\documentclass[sigconf]{acmart}

\usepackage{graphicx}
\usepackage{subcaption}
\usepackage{multirow}
\usepackage{enumitem}
\usepackage{tikz}
\usepackage{tabularx}

\usepackage[acronym]{glossaries}
\makeglossaries

\newacronym{AI}{AI}{artificial intelligence}
\newacronym{ANN}{ANN}{artificial neural network}
\newacronym{LLM}{LLM}{large language model}
\newacronym{NLP}{NLP}{natural language processing}
\newacronym{NLI}{NLI}{natural language inference}
\newacronym{RAM}{RAM}{random access memory}
\newacronym{VRAM}{VRAM}{video random access memory}
\newacronym{GPU}{GPU}{graphics processing unit}
\newacronym{KG}{KG}{knowledge graph}

\newacronym{CoT}{CoT}{Chain-of-Thought}
\newacronym{SC}{SC}{Self-Consistency}
\newacronym{CP}{CP}{Chat Protect}
\newacronym{ToT}{ToT}{Tree-of-Thoughts}
\newacronym{CoVe}{CoVe}{Chain-of-Verification}
\newacronym{MAD}{MAD}{Multiagent Debate}
\newacronym{DDGA}{DDGA}{DuckDuckGo Augmentation}
\newacronym{KGR}{KGR}{Knowledge Graph-based Retrofitting}
\newacronym{ReAct}{ReAct}{Reasoning and Acting}

\newacronym{GSM8K}{GSM8K}{Grade School Math 8K}
\newacronym{MMLU}{MMLU}{Massive Multitask Language Understanding}

\usetikzlibrary{shapes.geometric, arrows}
\tikzstyle{startstop} = [rectangle, rounded corners, minimum width=2cm, minimum height=0.6cm, text centered, draw=black, fill=red!30]
\tikzstyle{process} = [rectangle, minimum width=2cm, minimum height=0.6cm, text centered, draw=black, fill=orange!30]
\tikzstyle{decision} = [diamond, minimum width=2cm, minimum height=0.6cm, text centered, draw=black, fill=green!30, aspect=2]
\tikzstyle{dialogue} = [rectangle, rounded corners, minimum width=0.95\textwidth, minimum height=1cm, 
    text width=0.95\textwidth, text centered, draw=black, fill=blue!20, inner sep=10pt, font=\itshape]
\tikzstyle{arrow} = [thick,->,>=stealth]

\begin{document}

\title{Investigating the Role of Prompting and External Tools in Hallucination Rates of Large Language Models}

\author{Liam Barkley}
\email{24899518@sun.ac.za}
\author{Brink van der Merwe}
\email{abvdm@sun.ac.za}
\affiliation{
  \institution{Stellenbosch University}
  \city{Stellenbosch}
  \state{Western Cape}
  \country{South Africa}
}

\begin{abstract}
    \Glspl{LLM} are powerful computational models trained on extensive corpora of human-readable text, enabling them to perform general-purpose language understanding and generation. \glspl{LLM} have garnered significant attention in both industry and academia due to their exceptional performance across various \gls{NLP} tasks. Despite these successes, \glspl{LLM} often produce inaccuracies, commonly referred to as \emph{hallucinations}. Prompt engineering, the process of designing and formulating instructions for \glspl{LLM} to perform specific tasks, has emerged as a key approach to mitigating hallucinations. This paper provides a comprehensive empirical evaluation of different prompting strategies and frameworks aimed at reducing hallucinations in \glspl{LLM}. Various prompting techniques are applied to a broad set of benchmark datasets to assess the accuracy and hallucination rate of each method. Additionally, the paper investigates the influence of tool-calling agents---\glspl{LLM} augmented with external tools to enhance their capabilities beyond language generation---on hallucination rates in the same benchmarks. The findings demonstrate that the optimal prompting technique depends on the type of problem, and that simpler techniques often outperform more complex methods in reducing hallucinations. Furthermore, it is shown that \gls{LLM} agents can exhibit significantly higher hallucination rates due to the added complexity of external tool usage.
\end{abstract}


\settopmatter{printacmref=false}
\renewcommand\footnotetextcopyrightpermission[1]{}
\pagestyle{plain}

\maketitle

\section{Introduction}

\Glspl{LLM} are computational models capable of general-purpose language generation. \glspl{LLM} have become ubiquitous in recent years owing to their ability to perform a variety of \gls{NLP} tasks, such as language translation, comprehension, textual summarization, and question-answering. Moreover, \glspl{LLM} have shown promise in solving complex reasoning tasks such as writing code or answering mathematical problems \cite{minaee2024survey}. However, despite their successes, current \glspl{LLM} exhibit a concerning tendency to generate inaccurate or misleading information, often referred to as \emph{hallucinations}. Owing to their stochastic nature, \glspl{LLM} may produce plausible outputs that lack or contradict real-world evidence.

Hallucinations from \glspl{LLM} can produce fallacies, biases or misinformation, which is particularly concerning as they garner widespread use. One of the most accessible and widely used \glspl{LLM} to date, ChatGPT, owned by OpenAI, has approximately 185.5 million users \cite{mortensen2024chatgpt}. Furthermore, one of the world's leading instant messaging platforms, \texttt{WhatsApp}, introduced an \gls{LLM}-powered chatbot on the platform in April 2024, further facilitating the use of these models by the public. 
The growing popularity of \glspl{LLM}, coupled with a general lack of understanding, can lead to crucial inaccuracies, especially in political or medical contexts, with potentially serious consequences. Therefore, it is imperative to study the causes of hallucinations in \glspl{LLM} and develop methods to mitigate them.

\gls{LLM} prompts are the text-based input that allows users to interact with the model. It is usually a question or set of instructions that triggers a specific response or action in the \gls{LLM}. Prompt engineering is the nuanced practice of designing \gls{LLM} prompts to optimize the output of the \gls{LLM} so that it completes the desired task correctly. Prompting frameworks offer a systematic way to design prompts so that the \gls{LLM} elicits the desired output. While certain prompting frameworks are specifically designed for the detection and mitigation of hallucinations, general prompt engineering also facilitates the reduction of hallucinations since the goal is to optimize the correctness of \gls{LLM} responses.

\gls{LLM} agents are \gls{AI} systems designed to solve complex tasks. The main idea behind \gls{LLM} agents is to perform a sequence of actions using an \gls{LLM} as the central reasoning engine, that dictates which actions to take and in what order. These agents are often augmented with external tools to expand the capabilities of the model beyond language generation. At the core of these systems are entire prompting frameworks, often referred to as agent architectures, which dictate how many steps the model may take and what it can do at each step. Given that \gls{LLM} agents are frequently employed to generalize and extend the functionality of base models, it is important to investigate their potential impact on hallucinations

This paper provides a comprehensive empirical evaluation of various black-box approaches, such as prompt engineering and the use of \gls{LLM} agents, on the hallucination rates of different \gls{NLP} tasks. This focus is motivated by the fact that many state-of-the-art \glspl{LLM} are proprietary, and users typically do not have access to the internal workings of these models. Furthermore, hallucinations are highly context-dependent. For example, they could yield the generation of novel ideas in creative writing contexts. Therefore, this paper investigates techniques that are applicable across different contexts, regardless of the implementation details of the underlying model. The goal of this study is to offer valuable insights into the effectiveness and applicability of prompt engineering and different agent architectures, so that it may pave the way for more secure and reliable \gls{LLM} applications across a variety of domains.

\section{Background} \label{sec:background}

This section provides a summary of the concepts discussed in this paper. It gives a brief overview of LLMs, hallucinations, prompting techniques and LLM agents.

\subsection{Large Language Models}

The most powerful and capable \glspl{LLM} to date are transformer-based models, which are a specific type of neural network introduced by Google in 2017 \cite{Vaswani2017}. \Glspl{LLM} are based on layers of artificial neurons that process sequences of tokenized text such as words, sub-words, or individual characters. The transformer architecture is a specific type of neural network that utilizes an ``attention'' mechanism, which enables the model to selectively focus on different parts of the input data. This mechanism is particularly well-suited at capturing long-range dependencies and complex patterns in language data. Through an extensive training process, these models adjust the weights and biases between neurons to predict the most probable tokens following a specific input sequence, which yields the production of coherent and human-like text. Increasing the number of parameters of the model, enables the model to capture more intricate language patterns. However, this increase is subject to diminishing returns, data quality and other factors.

Running \glspl{LLM} locally is computationally expensive, since these models have to process vast amounts of data in real time by performing large-scale matrix multiplications. Significant amounts of \gls{RAM} and \gls{VRAM} are required, with \glspl{GPU} often recommended for their efficiency in handling parallel operations like matrix multiplications. For example, running an 8 billion parameter model would generally require a \gls{GPU} with at least 8GB of \gls{VRAM} to load the entire model onto the \gls{GPU}. Although it is possible to run \glspl{LLM} without a \gls{GPU}, it requires a significant amount of \gls{RAM} and generally results in slower inference times. Larger models, such as those exceeding 70 billion parameters, may require over 64GB of \gls{RAM} or 40GB of \gls{VRAM} to yield reasonable inference times.

\gls{LLM} quantization is a model compression technique used to reduce the size of \glspl{LLM} by converting the high-precision weights of the model to lower-precision weights. Model weights are often stored in a high-precision format, such as 32-bit floating point numbers. \gls{LLM} quantization tries to make the model more energy efficient by casting the weights to a lower precision, such as 8-bit integers, while maintaining the performance of the model. Quantization effectively reduces the memory and storage consumption of a model so that it utilizes less computational resources. The reduced model size results in faster inference times and less energy consumption, making quantization an essential technique to run \glspl{LLM} on smaller, less powerful devices.

The temperature of an \gls{LLM} is a hyperparameter that balances exploration and exploitation of the output generated by the model. When \glspl{LLM} generate new tokens, there are often a few candidates to choose from, each with varying probabilities of being selected. The temperature setting adjusts the probability distribution over the candidate tokens. Specifically, when the temperature is low (close to zero), the model exploits patterns it has learned in the training data, making the output more reliable and predictable. On the contrary, larger temperature values (usually close to or greater than one), facilitate exploration of the token generation space, by increasing the probability of selecting more diverse and unpredictable tokens. Therefore, lower temperatures result in more coherent and expected responses, whereas higher temperatures result in more creative and diverse responses.

\subsection{Hallucination Taxonomy}

The terminology used to describe hallucinations varies significantly across current literature on \gls{LLM} hallucinations. Many works have divided hallucinations into two categories, intrinsic and extrinsic hallucinations \cite{amatriain2024llmhallucinations, minaee2024survey, ji2023survey}. Intrinsic hallucinations generally refer to any output that has a direct contradiction to some source material, whereas extrinsic hallucinations are defined as any output that includes speculative content which is not based on the provided source material. However, this nomenclature is restricted to tasks that include source material, such as text summarizations or reading comprehension tasks.

Therefore, Huang \textit{et al.} \cite{huang2023survey} proposed a new taxonomy for hallucinations in \glspl{LLM} that better encapsulates the various types of \gls{NLP} tasks that can incur hallucinations. Their proposed taxonomy includes two general categories, factuality and faithfulness hallucinations. A factual hallucination is defined as any information generated by an \gls{LLM} that contradicts or is not supported by real-world knowledge. Factual hallucinations are divided into two subcategories, factual inconsistencies and factual fabrications. Factual inconsistencies are any facts that directly contradict real-world knowledge, whereas factual fabrications are any facts that are not supported, nor contradicted, by real-world knowledge.

Faithfulness hallucinations are \gls{LLM} responses that do not align with prompt instructions or any additional context. Faithfulness hallucinations can be divided into three categories: instruction inconsistencies, context inconsistencies, and logical inconsistencies. Instruction inconsistencies are any responses that do not align with the prompt instructions. Contextual inconsistencies contradict any additional context provided in the prompt, and logical inconsistencies are when the model contradicts itself within the same response.

\subsection{Prompt Engineering}

Prompt engineering is the art of constructing \gls{LLM} prompts that yield the most relevant and correct responses. \Gls{CoT} prompting is a strategy introduced by Wei \emph{et al.} \cite{wei2023chainofthought} where the \gls{LLM} has to elicit explicit reasoning for its response. This technique is particularly powerful for reasoning tasks such as mathematical problem-solving. It improves the reasoning capability of the \gls{LLM} by breaking the task into smaller steps to be solved.

\Gls{SC} is a technique proposed by Wang \emph{et al.} \cite{wang2023selfconsistency} that performs a majority vote based on several repeated \gls{LLM} calls. \gls{LLM} models are often encouraged to perform greedy decoding by biasing the \glspl{LLM} output to the safest and most predictable response. This is achieved by adjusting settings such as the temperature value of the \gls{LLM}. The \Gls{SC} approach aims to balance creativity with accuracy by sampling from diverse \gls{LLM} responses and performing a majority vote over the sampled answers.

Similarly, \Gls{ToT} is a prompting strategy developed by Yao \emph{et al.} \cite{yao2023tree} for deliberate problem-solving. This strategy entails sampling different reasoning paths for a particular problem. It involves subdividing the problem into smaller steps and generating solutions for each step. At each step, a separate prompt is used to evaluate and vote for the best path of reasoning. This process is continued until the final step is completed. Contrary to \gls{SC}, \gls{ToT} emphasizes the steps used to solve a specific problem as opposed to a majority vote of the final solution to the problem.

\emph{Reflection} is a simple prompting strategy that is based on the fact that \glspl{LLM}, like humans, often do not get things right on their first try. This strategy contains two \glspl{LLM}, a generator and a reflector. First, the generator attempts to respond to the query of the user, then the reflector is prompted to provide constructive criticism on the response from the generator. The critique and feedback are then sent back to the generator to produce a new response based on the feedback. This process is repeated for a desired number of iterations.

\subsection{Frameworks to Mitigate Hallucinations}

Many frameworks have been proposed to mitigate hallucinations in \glspl{LLM}. M\"undler \textit{et al.} \cite{mundler2024} proposed a framework, known as \gls{CP}, to reduce hallucinations based on contradictory responses. Dowden \cite{dowden1993} stated that given any two contradictory responses that describe the same subject, at least one of the claims are guaranteed to be false. This forms the basis of the \gls{CP} approach, which entails a three-stage pipeline to invoke, detect and remove contradictory claims from \gls{LLM} responses. The approach uses an analysing \gls{LLM} to detect and remove false claims by a generating \gls{LLM}. During the invocation stage, the algorithm extracts contexts from each sentence in the generating \glspl{LLM} response. Then, the generating \gls{LLM} is queried based on the restricted contexts to yield a new response for each independent context. Finally, the analysing \gls{LLM} compares each set of responses to detect and remove contradictory statements from the output of the generating \gls{LLM}.

Furthermore, Guan \textit{et al.} \cite{guan2023mitigatinglargelanguagemodel} suggested an approach to mitigate factual hallucinations by grounding an \gls{LLM} with information from an external \gls{KG}. A \gls{KG} is a structured representation of real-world entities and their relationships. Nodes in the graph, called entities, represents real-world objects or concepts such as people, places or items. Related entities are connected by edges in the graph. Each edge in the graph contains information about the relationship between the two connected entities. Therefore, \glspl{KG} store information about the world in a format that makes it simple and efficient to query general facts about the world. The proposed approach, called \gls{KGR}, enables autonomous \gls{KG} retrieval by using an \gls{LLM} to extract entities from an initial draft response and searching the \gls{KG} for properties chosen by the \gls{LLM}. The information from the \gls{KG} is then added as additional context so that the \gls{LLM} can refine its initial response. This enables the \gls{LLM} to ground its final response in external knowledge from the \gls{KG}, in order to decrease the number of factual hallucinations. 

Inspired by the work of Minsky's \emph{The Society of Mind} \cite{minsky1988}, Du \textit{et al.} \cite{du2023} put forward a framework to reduce hallucinations based on the concept of interaction between cognitive components. The \gls{MAD} framework is based on a form of collective intelligence where multiple \glspl{LLM} work together to procure a response. The idea is that contradictions, and therefore hallucinations, can be reduced by having multiple \glspl{LLM} with diverse responses debate about their reasoning and obtain a convergent solution. The solution that the \glspl{LLM} converge to is more likely to contain factual information, according to Minsky \cite{minsky1988}. This approach follows a three-step process. First, each \gls{LLM} is prompted to generate an independent response. Secondly, the debate is initiated by prompting each \gls{LLM} to revise their response given the responses of the other \glspl{LLM}. This step is repeated for a fixed number of iterations. Finally, an \gls{LLM} is prompted to combine the final responses from each \gls{LLM} to produce a single response. 

Finally, Dhuliawala \textit{et al.} \cite{dhuliawala2023} introduced \gls{CoVe}, a four-step process to reduce hallucinations using a set of verification questions. First, the \gls{LLM} generates an initial response. Secondly, it generates verification questions, based on the query and the initial response, that can be used to verify key facts in the base response. The verification questions are then answered independently and evaluated by the \gls{LLM} to identify contradictions between the independent answers and the base response. Finally, the \gls{LLM} removes contradicting claims from the original response by taking into account the independent answers from the verification questions. Three validation methods are proposed: the \emph{joint} method, which combines question generation and answering in one query, the \emph{2-Step} approach which separates these into two independent queries, and the \emph{factored} strategy where each question is answered with an independent query. The \emph{factored} approach is the most computationally expensive approach, but has the lowest likelihood of carrying over hallucinations from the base response.

\subsection{Agents}

Agentic systems are \gls{LLM}-based applications where the control flow is determined by an \gls{LLM}. In agent-based systems, the agent architecture governs the interaction between \glspl{LLM}, external systems, and the control flow of the system \footnote{Agent architectures: \url{https://langchain-ai.github.io/langgraph/concepts/agentic_concepts/}.}. Different architectures yield varying degrees of control by the \gls{LLM}. The simplest is a chain architecture, where tasks are solved sequentially with a pre-determined sequence of \gls{LLM} calls. Router architectures offer a more dynamic system, where the \gls{LLM} governs the flow of the system by selecting from a set of pre-defined chains. A more sophisticated architecture is that of general tool-calling agents, where the \gls{LLM} is responsible for multistep decision-making and tool calls. \Gls{ReAct} \cite{react2023} is a popular general-purpose architecture that interleaves reasoning with task-specific actions and incorporates the following three modules:
\begin{itemize}
    \item \textbf{Tools}: External tools available to the model.
    \item \textbf{Memory}: Retain information from prior steps.
    \item \textbf{Planning}: Dictate the steps taken to accomplish a task.
\end{itemize}

Tools enable sufficiently trained \glspl{LLM} to access external systems for tasks such as arbitrary code execution, looking up information online or executing specialized actions. Tool binding involves giving a model awareness of the tools it has available to it and specifying the required tool calling schema. The conventions for formatting tool calls vary between different \gls{LLM} providers; for example, OpenAI uses JSON, whereas other providers use parsed content blocks. In the \gls{ReAct} architecture, the planning component uses a while-loop that consists of a thought, an action, and an observation. The thought dictates which tool to call, the action includes the specific tool calling schema with the desired arguments, and the output contains the result of the tool invocation. The \gls{LLM} terminates the loop once it has completed its goal. 

\section{Methodology and Experimental Design} \label{sec:methodology_and_experimental_design}

This section details the \gls{LLM}-based system, including the models and libraries used, the implementation of the various prompting strategies and agents, and the process for evaluating and comparing these approaches on different benchmarks.

\subsection{Implementation}

The different prompting strategies, frameworks, and agents were implemented using Python, LangChain, and Ollama. LangChain is an open-source Python framework designed to build \gls{LLM}-powered applications. LangChain offers an extensive range of components to design, develop, and integrate existing \glspl{LLM} into Python applications, which made it well-suited for implementing and testing the different prompting strategies and agents. Ollama is an open-source software platform to run \glspl{LLM} on a local machine. LangChain provides functions for interacting with models running on Ollama. All the models were hosted locally on an Nvidia Geforce RTX 2080 \gls{GPU} with 8GB of dedicated \gls{VRAM}. 

The prompting strategies and frameworks were tested using the Meta-Llama-3-8B-Instruct-Q6\_K model, which has 8 billion parameters and 6-bit quantization. The model was tested with temperature values of 0.2, 0.5, and 0.8 to discern the impact of different values on mitigating hallucinations with the various prompting strategies. Furthermore, since the  Meta-Llama-3-8B-Instruct-Q6\_K model does not support tool use, the agent architectures were implemented using the Meta-Llama-3.1-8B model with 8 billion parameters and no quantization. Owing to time constraints, the agents were all tested with a temperature value of 0.5.

\subsection{Prompting Techniques}

The \gls{CoT} strategy \footnote{For a detailed description of the prompts used, refer to Appendix \ref{apx:prompts}.} was implemented for reasoning-based \gls{NLP} tasks by parsing both the final answer of the model and the sequence of steps used to solve the problem. This approach required the model to explicitly include and format its reasoning process, thereby dividing the problem into smaller, more manageable parts. However, \glspl{LLM} may not always adhere to the specific output instructions, which can result in parsing errors. This was mitigated by allowing multiple attempts, up to a specified tolerance number. Invalid responses were discarded since they could not be parsed into a structured format. If the number of attempts exceeded the tolerance threshold, an error was raised and the query was invalid. The \gls{ToT} approach involved sampling from the \gls{CoT} prompt multiple times. Next, the \gls{LLM} was prompted to select the most accurate solution based on the parsed reasoning steps, with the final answer being obtained from the best-voted sample. A control strategy was implemented to compare each method against the base model. The control strategy only required the formatted answer, allowing the base model to decide when to include reasoning steps. The \gls{SC} strategy used repeated sampling of the control strategy to select an answer based on a majority vote, while the \gls{SC}-\gls{CoT} strategy involved a majority vote over sampled \gls{CoT} responses. The \gls{CP} strategy involved sampling from the control strategy several times, where any contradictory answers resulted in the model refraining from answering. Owing to the amount of computational power available, and to prevent ties in the majority vote, the number of samples for the \gls{SC}, \gls{SC}-\gls{CoT}, \gls{ToT} and \gls{CP} strategies was chosen to be five.

The \gls{MAD} framework was implemented using a conversation buffer from LangChain, so that the agents could recall previous iterations of the debate. It is a form of short-term memory that automatically includes previous user queries and responses in the prompt when new queries are made. For simplicity, the \gls{MAD} implementation only utilized two debating \glspl{LLM}. The debate was terminated whenever the two \glspl{LLM} agreed on a solution or after reaching a maximum number of ten iterations, whereby the solution would be taken from the final answer of the first \gls{LLM}. Similarly, the reflection strategy consisted of a single iteration of explicit feedback. After generating an initial response, the reflector \gls{LLM} acted as a teacher grading an exam submission by offering constructive criticism on the initial response. This feedback was then used as additional context for the \gls{LLM} to enhance its final response.

Two variants of the \gls{CoVe} framework were implemented and tested. The first variant, called \gls{CoVe}-1, was designed for answering basic general-knowledge questions. This approach involved generating a single verification question based on an initial unformatted response. The verification question was then answered independently, and a fourth \gls{LLM} query was done to determine whether the answer to the verification question contradicted the initial response in any way. If a contradiction was detected, the model would refrain from answering the question, otherwise the model would generate a formatted response for the original query. The second \gls{CoVe} variant, \gls{CoVe}-2, was developed for multiple choice. This approach involved sampling an initial multiple choice answer, generating a second response without giving the options, and then checking if the second response matched the original choice. If they aligned, the original choice was returned, otherwise the model refrained from answering.

The \gls{KGR} implementation utilized the Wikidata \gls{KG} \cite{vrandecic2014wikidata}, which is a freely available and collaboratively constructed \gls{KG}. After generating an initial answer, the \gls{LLM} extracted a relevant entity based on the question and attempted answer. Next, the \gls{LLM} selected an appropriate property of the entity to retrieve from the \gls{KG}. Finally, the retrieved information was added as additional context to generate a final response to the original query. The \gls{DDGA} strategy was developed to compare approaches that ground the model with external information. \texttt{DuckDuckGoSearchRun} is a Python search engine that can retrieve snippets of information from the internet based on a search query. The \gls{DDGA} approach involved the retrieval of external information by performing a DuckDuckGo search of the user's input. The retrieved information was added as additional context to the \gls{LLM} prompt to ground the model in external information.

\subsection{Agent Architectures}

To investigate the effect of agents on hallucination rates in different \gls{NLP} tasks, two agent architectures were implemented and compared with a control agent. The first agent used a simple chain architecture that consisted of two \gls{LLM} queries. The first query generated a list of tool calls, and the second used the outputs from these tools to provide a final answer. The LangChain \texttt{bind\_tools} function was used to integrate tools with the model. The second agent utilized a general-purpose \gls{ReAct} architecture. This was achieved by using the \texttt{create\_tool\_calling\_agent} and \texttt{AgentExecutor} functions from LangChain. Each agent was equipped with a combination of three tools: Wikipedia, DuckDuckGo, and Riza, a Python interpreter. The Wikipedia tool handled queries about people, places, or items, while DuckDuckGo enabled general internet searches for up-to-date information. Riza allowed for the execution of arbitrary Python code in a secure sandbox environment, to avoid potential issues from agents generating dangerous or non-terminating code. Additionally, a third \gls{ReAct} agent, \gls{ReAct}-DDG, was introduced, limited to the DuckDuckGo search tool.

\subsection{Benchmarks}

The following benchmarks were used to evaluate each of the algorithms implemented. First, is the \gls{GSM8K} benchmark \cite{gsm8k2021dataset}, which contains a collection of mathematical word problems that require a sequence of logical steps to solve. The test set contains 1319 questions, each with a single numerical solution. This benchmark evaluated the extent to which each strategy could reduce logical hallucinations. The prompting strategies that were tested on this benchmark were the \gls{CoT}, \gls{SC}, \gls{SC}-\gls{CoT}, \gls{ToT} and \gls{MAD} strategies, which are all aimed at improving reasoning capabilities. Furthermore, owing to time constraints, the agent architectures were only evaluated on the first 1000 questions.

Secondly, was the TriviaQA benchmark \cite{triviaqa2017dataset}, which consists of reading comprehension and high quality trivia questions. The TriviaQA dataset was used to determine each strategy's ability to mitigate factual inconsistencies. In particular, each selected strategy was assessed on the first 1000 trivia questions from the validation set. The \gls{SC}, \gls{CP}, \gls{KGR}, \gls{CoVe}-1, \gls{MAD} and \gls{DDGA} strategies, that all aim to mitigate factual inconsistencies, were evaluated on the TriviaQA benchmark. The \gls{ReAct}-DDG agent was only evaluated on the TriviaQA benchmark to discern the impact of having less tools compared to the \gls{ReAct} and chain architectures on this benchmark.

The final benchmark was the \gls{MMLU} dataset, which consists of general-knowledge multiple choice questions, spanning 57 different subjects. The selected strategies were assessed on 1000 questions in total, that included approximately 17 questions per subject. The strategies applied to this benchmark set were the \gls{SC}, \gls{CP}, \gls{MAD}, reflection and \gls{CoVe}-2 strategies, as well as the chain and \gls{ReAct} agents.

\subsection{Evaluation Metrics}

For each benchmark, the number of correct answers, the number of hallucinated answers and  the accuracy was computed over a number of independent runs. Due to time constraints and limited computational resources, each strategy was run three times per benchmark. This was done since the output of an \gls{LLM} is stochastic, which makes it important to obtain an indication of average performance. The Top-N accuracy was used to investigate the influence of temperature on strategies with repeated sampling. This metric indicates the percentage of times that the correct answer appeared in N sampled answers. The performance of the base \gls{LLM} was evaluated according to each metric as a control method. The results were tabulated to determine which methods yield the greatest reduction in hallucinations. Additionally, the table includes the average number of prompts per strategy to indicate the average computational cost for each method. Owing to the limited number of runs, no statistical tests were conducted since the power of the statistical results would be very low.

\section{Results} \label{sec:results}

This section presents the results of each algorithm on the benchmark datasets. The first three parts evaluate the performance of prompting techniques over the different benchmarks, and the final part of this section discusses and analyses the results of the agents.

\subsection{GSM8K Results}

Table \ref{tab:gsm8k_results} indicates the means of the \gls{CoT}, \gls{SC}, \gls{SC}-\gls{CoT}, \gls{ToT} and \gls{MAD} strategies for different temperature values on the \gls{GSM8K} dataset \cite{gsm8k2021dataset} over the independent runs. The best value for each performance metric is given in bold. It is evident from the results that the \gls{SC} strategy, with a temperature value of 0.8, had the best overall performance on the benchmark by achieving the highest accuracy and least number of hallucinated answers on average. The \gls{SC} strategy achieved the best balance between accuracy and creativity amongst all the strategies on the \gls{GSM8K} benchmark.

\begin{table}[htbp]
\centering
\caption{Average performance of different prompting strategies, for various temperatures, on the \gls{GSM8K} benchmark.}
\begin{tabularx}{\columnwidth}{Xcccc}
\toprule
\textbf{Strategy} & \textbf{Cost} & \textbf{Hallucinated} & \textbf{Correct} & \textbf{Accuracy (\%)} \\
\midrule
\multicolumn{5}{c}{\textbf{Temperature 0.2}} \\
\midrule
Control & 1.00 & 288.33 & 1030.67 & 78.14 \\
\gls{CoT} & 1.00 & 326.33 & 990.00 & 75.21 \\
\gls{SC} & 5.00 & 229.34 & 1088.00 & 82.59 \\
\mbox{\gls{SC}-\gls{CoT}} & 5.00 & 234.00 & 1078.00 & 82.16 \\
\gls{ToT} & 6.00  & 273.00 & 1037.67 & 79.17 \\
\gls{MAD} & 3.58 & 271.00 & 1030.67 & 79.18 \\
\midrule
\multicolumn{5}{c}{\textbf{Temperature 0.5}} \\
\midrule
Control & 1.00 & 302.33 & 1016.67 & 77.08 \\
\gls{CoT} & 1.00 & 338.33 & 980.67 & 74.34 \\
\gls{SC} & 5.00 & 213.66 & 1105.33 & 83.80 \\
\mbox{\gls{SC}-\gls{CoT}} & 5.00 & 212.33 & 1105.33 & 83.89 \\
\gls{ToT} & 6.00 & 288.67 & 1029.67 & 78.10 \\
\gls{MAD} & 3.52 & 266.67 & 1042.67 & 79.63 \\
\midrule
\multicolumn{5}{c}{\textbf{Temperature 0.8}} \\
\midrule
Control & 1.00 & 324.33 & 994.67 & 75.41 \\
\gls{CoT} & 1.00 & 381.33 & 937.33 & 71.08 \\
\gls{SC} & 5.00 & \textbf{199.33} & \textbf{1119.67} & \textbf{84.89} \\
\mbox{\gls{SC}-\gls{CoT}} & 5.00 & 230.34 & 1087.67 & 82.52 \\
\gls{ToT} & 6.00 & 317.33 & 1000.00 & 75.91 \\
\gls{MAD} & 3.52 & 270.00 & 1040.00 & 79.39 \\
\bottomrule
\end{tabularx}
\label{tab:gsm8k_results}
\end{table}

Table \ref{tab:gsm8k_results} shows that the \gls{SC} and \gls{SC}-\gls{CoT} strategies performed relatively similar on average, and both outperformed the control strategy. The repeated sampling enabled these strategies to elicit different ways of solving each mathematical problem. Higher temperature values yielded more diverse and creative answers, which increased the risk of hallucinations and inaccurate responses. This is evident by the fact that the performance of the control method deteriorated with an increase in temperature. Therefore, the repeated sampling of the \gls{SC} and \gls{SC}-\gls{CoT} approaches was able to counteract hallucinations by selecting the answer that appeared the most frequently. Since mathematics requires a certain degree of creativity as well as accurate reasoning, these two strategies struck an excellent balance between creative problem-solving and accurate reasoning.

\begin{figure}[htbp]
\centering
\begin{subfigure}{0.45\textwidth}
  \centering
  \includegraphics[width=\linewidth]{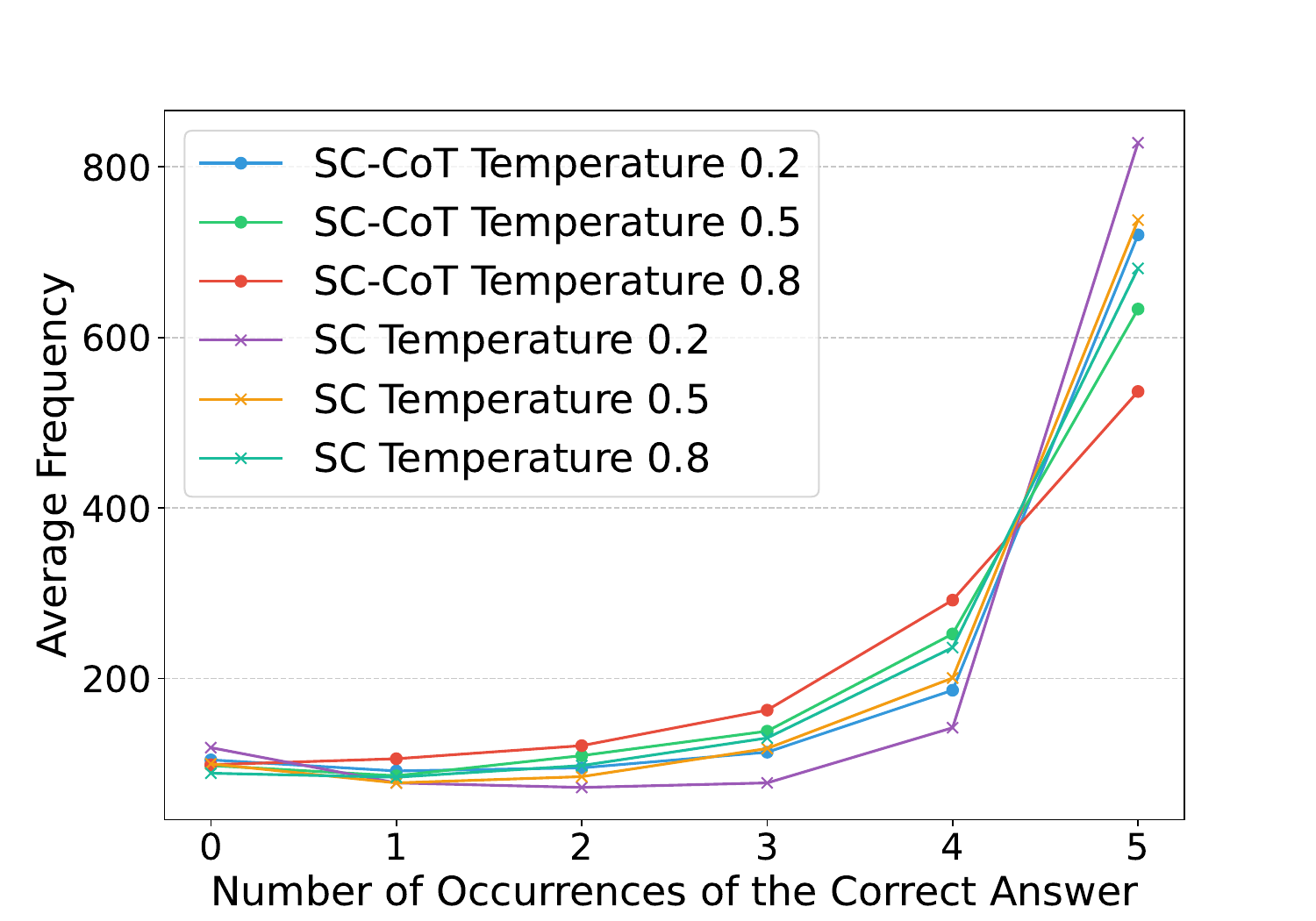}
\end{subfigure}
\caption{Average frequency over the number of correctly sampled responses per question for the \gls{SC} and \gls{SC}-\gls{CoT} strategies over the \gls{GSM8K} benchmark.}
\Description{A line graph showing the average frequency gradually increasing for each of the listed strategies.}
\label{fig:gsm8k_occurences}
\end{figure}

Figure \ref{fig:gsm8k_occurences} shows the average frequencies of how many times the correct answer appeared in the five sampled responses over the \gls{GSM8K} benchmark for the \gls{SC} and \gls{SC}-\gls{CoT} strategies. It is clear that the lower temperatures exhibited more consistent results. This is indicated by the fact that for both \gls{SC} and \gls{SC}-\gls{CoT}, a temperature value of 0.2 yielded high frequencies for having all five samples be correct and very low occurrences for only having one to four correctly sampled responses. On the contrary, the higher temperatures yielded more diverse responses, since the average frequency of correct occurrences is distributed more across the one to four range.

\begin{figure}[htbp]
\centering
\begin{subfigure}{0.45\textwidth}
  \centering
  \includegraphics[width=\linewidth]{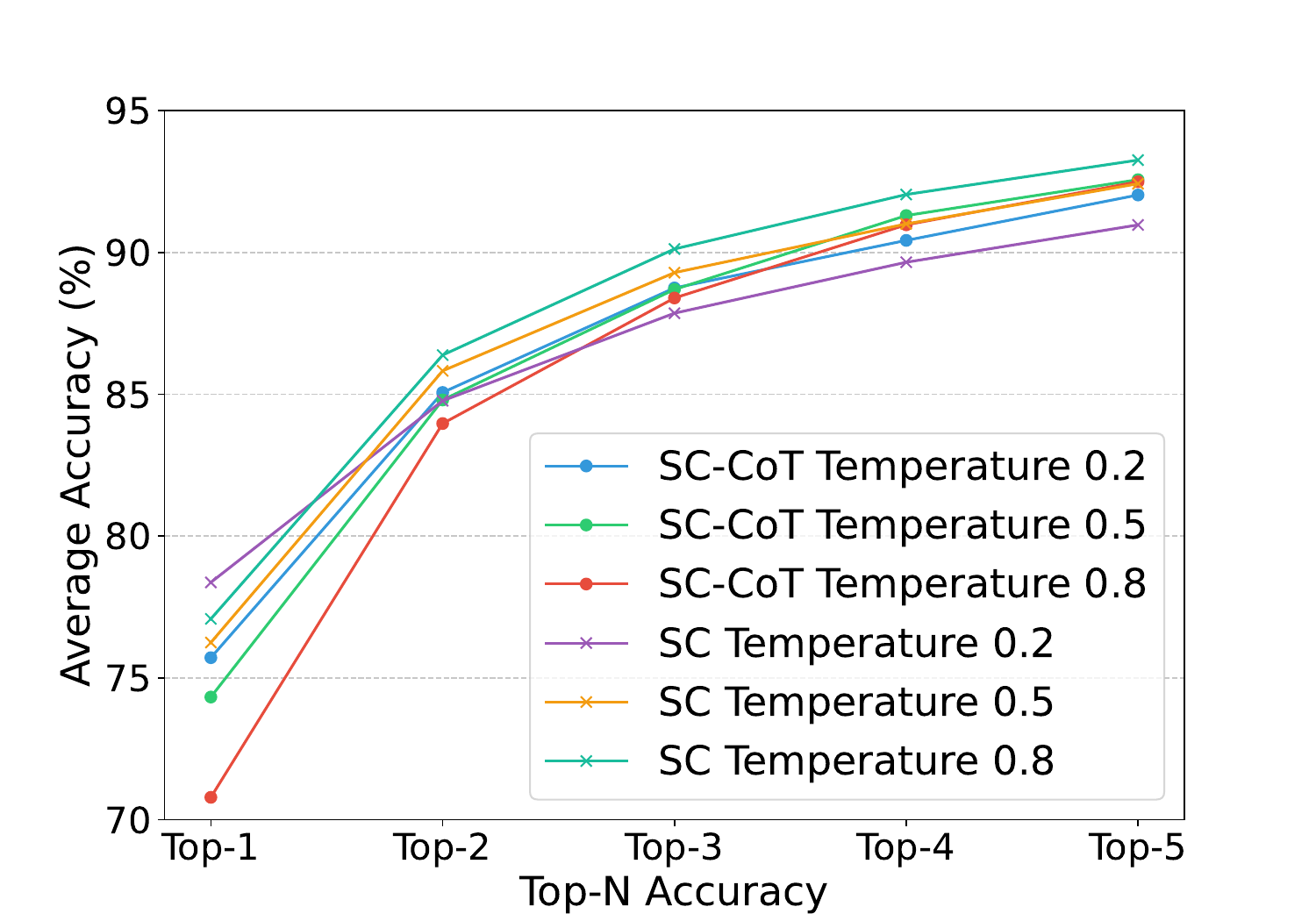}
\end{subfigure}%
\caption{Average Top-1 to Top-5 accuracy for the \gls{SC} and \gls{SC}-\gls{CoT} approaches on the \gls{GSM8K} benchmark.}
\Description{A line graph showing the average accuracy gradually increasing for each of the listed strategies.}
\label{fig:gsm8k_top_n}
\end{figure}

Figure \ref{fig:gsm8k_top_n} depicts the average Top-1 to Top-5 accuracy across different temperatures for the \gls{SC} and \gls{SC}-\gls{CoT} strategies. Figure \ref{fig:gsm8k_top_n} shows that a temperature of 0.2 led to the highest Top-1 accuracy for both \gls{SC} and \gls{SC}-\gls{CoT}, respectively. The lowest temperature value followed the safest, most correct reasoning paths, which achieved the highest Top-1 accuracies. On the contrary, the higher temperature values, of 0.5 and 0.8, achieved slightly better values for the Top-5 accuracy. The increased degree of randomness with the higher temperature values increased the probability of sampling the correct answer in at least one of the five sampled responses. 

An interesting observation is that the control strategy outperformed the \gls{CoT} strategy. One possibility for this is that the increased complexity of outputting explicitly formatted reasoning steps led to worse overall performance. This is supported by Figure \ref{fig:gsm8k_top_n} since the \gls{SC} approach achieved a higher Top-1 accuracy than \gls{SC}-\gls{CoT} for all the different temperature values, respectively. Furthermore, the \gls{ToT} and \gls{MAD} approaches only led to minor improvements over the control strategy. Therefore, these results suggest that the \gls{SC} sampling approach yielded the best results for mathematical reasoning and reducing logical hallucinations. 

\subsection{TriviaQA Results}

Table \ref{tab:triviaqa_results} depicts the average performance of the \gls{KGR}, \gls{CoVe}-1, \gls{MAD}, \gls{SC}, \gls{CP} and \gls{DDGA} strategies against the control method over the TriviaQA benchmark \cite{triviaqa2017dataset}. This table shows that the \gls{CP} strategy obtained the highest accuracy over all the strategies by greatly sacrificing the number of questions answered. Since the \gls{CP} approach refrained from answering questions where the sampled responses contained contradictory answers, it greatly reduced the number of hallucinations. As the temperature increased, the number of hallucinations decreased and the accuracy increased. Again, this is because the higher temperature values yielded more diverse responses and consequently more contradictory responses. Therefore, the temperature value dictated a trade-off between the number of hallucinations and the number of questions that the \gls{CP} approach answered. 

\begin{table}[htbp]
\centering
\caption{Average performance of different prompting strategies, for various temperatures, on the TriviaQA benchmark.}
\begin{tabularx}{\columnwidth}{Xcccccc}
\toprule
\textbf{Strategy} & \textbf{Cost} & \textbf{Graded} & \textbf{Halluc.} & \textbf{Correct} & \textbf{Acc. (\%)} \\
\midrule
\multicolumn{6}{c}{\textbf{Temperature 0.2}} \\
\midrule
Control & 1.00 & 1000.00 & 383.33 & 616.67 & 61.67 \\
\gls{KGR} & 4.00 & 873.00 & 327.33 & 546.00 & 62.54 \\
\gls{CoVe}-1 & 5.00 & 697.67 & 213.67 & 484.00 & 69.38 \\
\gls{MAD} & 3.20 & 999.00 & 365.33 & 633.67 & 63.43 \\
\gls{SC} & 5.00 & 1000.00 & 381.33 & 618.67 & 61.87 \\
\gls{CP} & 5.00 & 819.33 & 245.67 & 573.67 & 70.02 \\
\gls{DDGA} & 1.00 & 1000.00 & 323.67 & \textbf{676.33} & 67.63 \\
\midrule
\multicolumn{6}{c}{\textbf{Temperature 0.5}} \\
\midrule
Control & 1.00 & 1000.00 & 396.33 & 603.67 & 60.37 \\
\gls{KGR} & 4.00 & 864.00 & 341.00 & 523.00 & 60.53 \\
\gls{CoVe}-1 & 5.00 & 678.00 & 212.00 & 466.00 & 68.73 \\
\gls{MAD} & 3.65 & 999.00 & 365.00 & 634.00 & 63.46 \\
\gls{SC} & 5.00 & 1000.00 & 381.33 & 618.67 & 61.87 \\
\gls{CP} & 5.00 & 650.33 & 139.33 & 511.00 & 78.59 \\
\gls{DDGA} & 1.00 & 1000.00 & 325.33 & 674.67 & 67.47 \\
\midrule
\multicolumn{6}{c}{\textbf{Temperature 0.8}} \\
\midrule
Control & 1.00 & 1000.00 & 399.67 & 600.33 & 60.03 \\
\gls{KGR} & 4.00 & 831.67 & 329.33 & 502.33 & 60.40 \\
\gls{CoVe}-1 & 5.00 & 664.67 & 206.00 & 458.67 & 69.01 \\
\gls{MAD} & 3.97 & 999.33 & 376.67 & 622.67 & 62.31 \\
\gls{SC} & 5.00 & 1000.00 & 383.67 & 616.33 & 61.63 \\
\gls{CP} & 5.00 & 571.00 & \textbf{100.00} & 471.00 & \textbf{82.49} \\
\gls{DDGA} & 1.00 & 1000.00 & 341.33 & 658.67 & 65.87 \\
\bottomrule
\end{tabularx}
\label{tab:triviaqa_results}
\end{table}

Similarly to \gls{CP}, the \gls{CoVe}-1 approach also refrained from answering questions where the algorithm detected contradictions in the model. These contradictions provide an indication of the model's confidence in answering a specific question. If the model never contradicts itself, it is a good indication that the training of the model enabled it to answer the corresponding question. If the model cannot give consistent output on a query, there is a good probability that a hallucination has occurred. Despite the \gls{CoVe}-1 strategy outperforming the control method in terms of accuracy, it suffered from a large reduction in the number of correctly answered questions. For a temperature value of 0.2, the \gls{CP} strategy achieved a higher accuracy than the control method while maintaining a similar number of correct answers. On the contrary, \gls{CoVe}-1 approach yielded a higher accuracy but also refrained from answering questions the model was capable of answering, as indicated by the reduction in the number of correct answers compared to the control method. 

\begin{figure}[htbp]
\centering
\begin{subfigure}{0.45\textwidth}
  \centering
  \includegraphics[width=\linewidth]{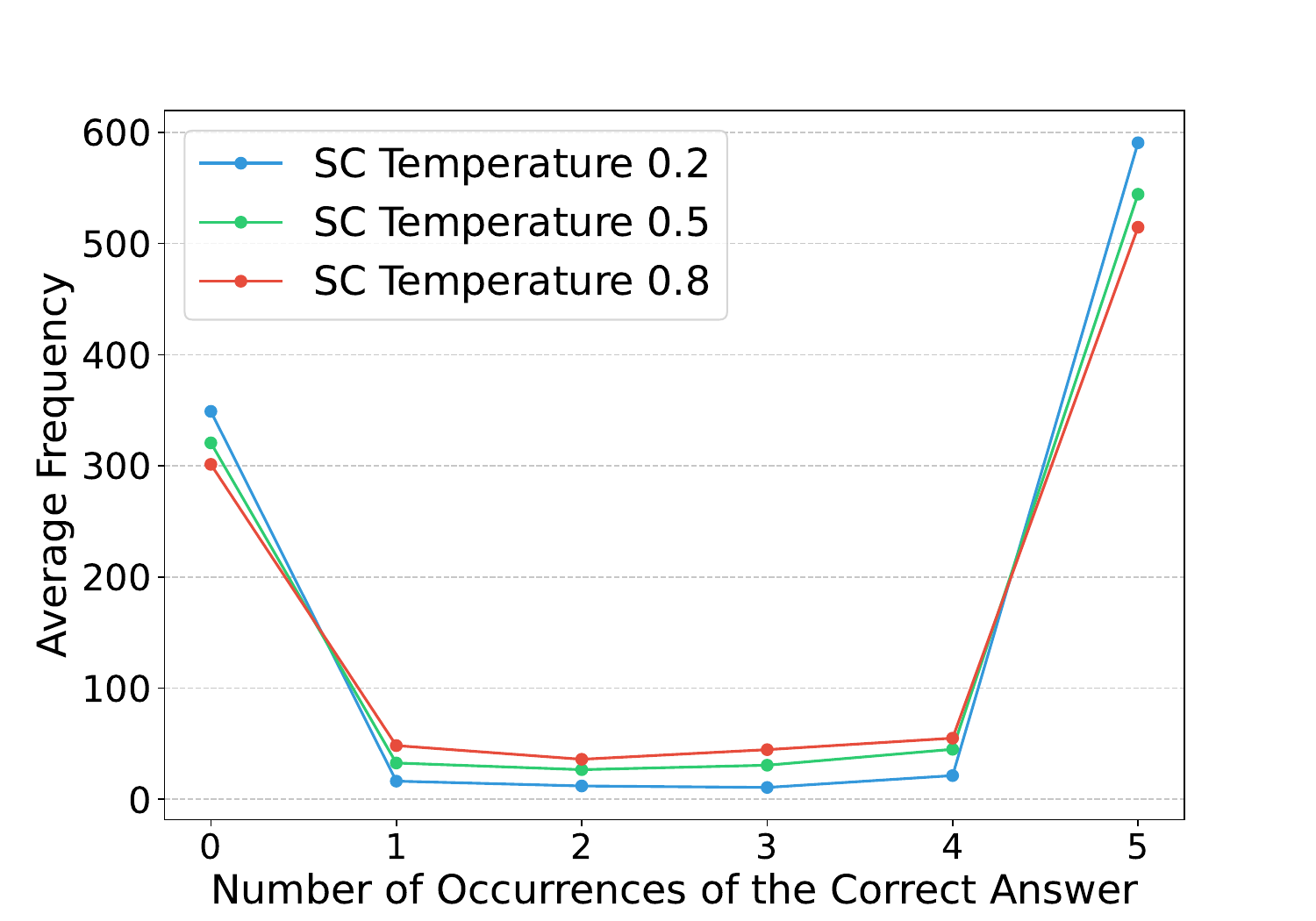}
\end{subfigure}
\caption{Average frequency over the number of correctly sampled responses per question for the \gls{SC} approach over the TriviaQA benchmark.}
\Description{A line graph showing the average frequency peaking at 0 occurences and 5 occurences.}
\label{fig:triviaqa_occurences}
\end{figure}

Figure \ref{fig:triviaqa_occurences} shows the average frequencies for the number of correctly sampled responses of \gls{SC} over the TriviaQA dataset. Figure \ref{fig:triviaqa_occurences} exhibits different characteristics to Figure \ref{fig:gsm8k_occurences} from the \gls{GSM8K} dataset. It is evident from Figure \ref{fig:triviaqa_occurences} that for most of the TriviaQA questions, the highest frequency of correctly sampled answers was either zero or five. This shows that on average, the model either got all or none of the sampled responses correct. Contrary to the \gls{GSM8K} dataset where the model had to obtain the correct answer via mathematical reasoning, the TriviaQA benchmark exhibited different sampling characteristics since there were no reasoning steps involved.

\begin{figure}[htbp]
\centering
\begin{subfigure}{0.45\textwidth}
  \centering
  \includegraphics[width=\linewidth]{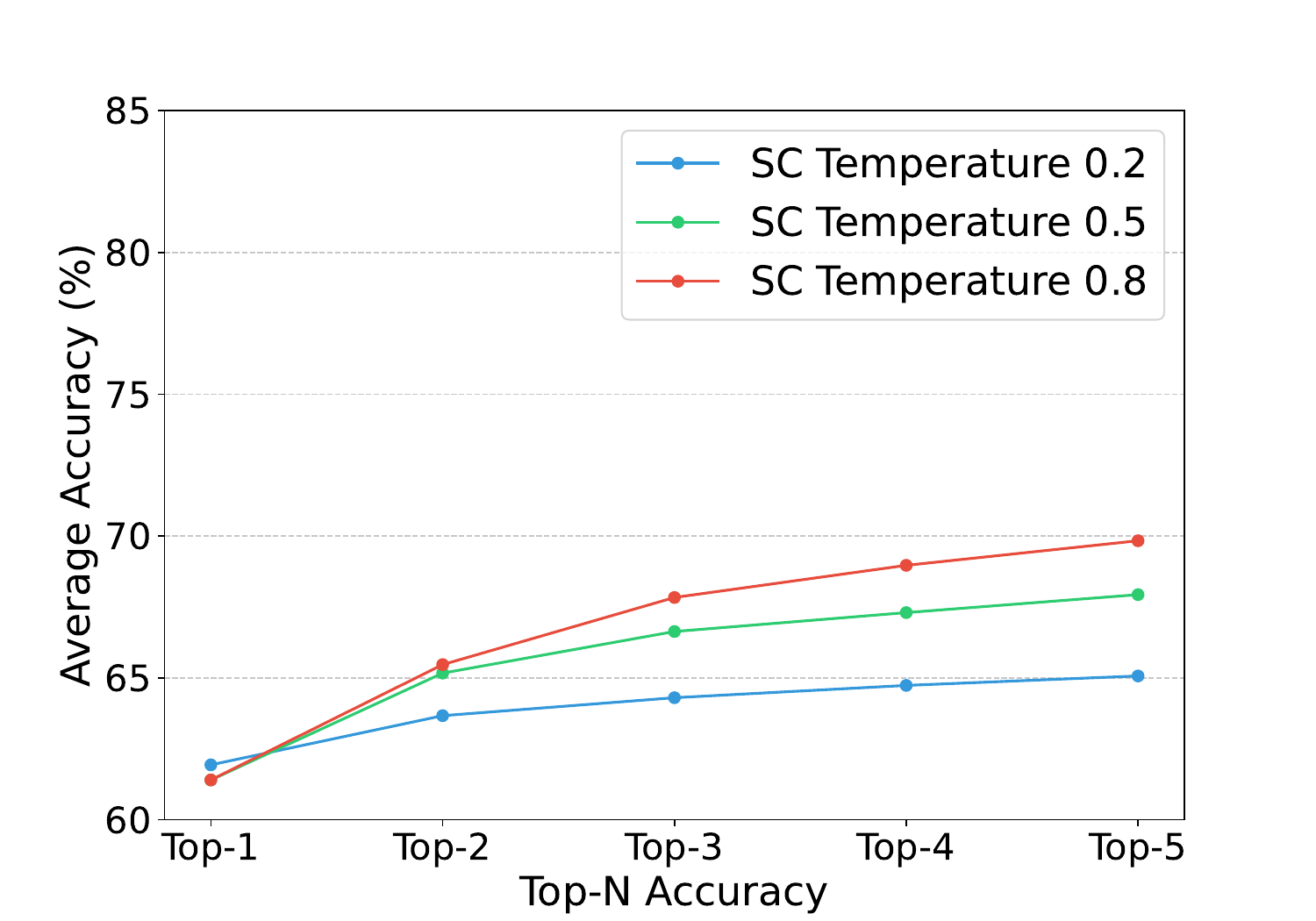}
\end{subfigure}%
\caption{Average Top-1 to Top-5 accuracy for the \gls{SC} strategy on the TriviaQA benchmark.}
\Description{A line graph showing the average accuracy gradually increasing for each of the listed strategies.}
\label{fig:triviaqa_top_n}
\end{figure}

Figure \ref{fig:triviaqa_top_n} shows the Top-1 to Top-5 accuracies for the \gls{SC} algorithm on the TriviaQA dataset. Contrary to Table \ref{tab:gsm8k_results}, increasing the temperature did not garner any significant improvements in the \gls{SC} approach. This is supported by the fact that both the \gls{SC} approach and the control strategy in Table \ref{tab:triviaqa_results} performed relatively similarly on average for all three temperatures, respectively. Additionally, the \gls{MAD} approach also performed relatively similar to the control strategy, with minor improvements for low-to-medium temperatures. This was expected for the TriviaQA dataset, since there were no reasoning steps involved. 

Moreover, it is evident from Table \ref{tab:triviaqa_results} that the \gls{DDGA} strategy managed to overcome the knowledge-boundary of the base model by grounding it with additional information from DuckDuckGo. The \gls{DDGA} approach obtained the highest number of correct answers out of all the strategies, including the control method. This suggests that the additional context provided by DuckDuckGo enabled the model to answer questions that it was not explicitly trained on. On the contrary, the \gls{KGR} approach exhibited much worse performance than the \gls{DDGA} approach. It was discovered during testing that the model often struggled to obtain relevant information from the Wikidata \gls{KG}. Consequently, the \gls{KGR} approach refrained from answering questions because it could not find any entities or properties relevant to the question. This implicitly resulted in fewer hallucinations compared to the control method. However, the \gls{KGR} approach could not extract meaningful information from the \gls{KG} and as a result the number of correct answers did not increase as it did for the \gls{DDGA} strategy.

The results of this benchmark suggest that the best strategy to employ for answering general-knowledge based questions is the \gls{CP} strategy. Furthermore, since the trade-off between the number of questions answered and the number of hallucinations can be controlled by the temperature parameter of the \gls{CP} approach, it allows for great versatility in the type of domain it is employed in. In scenarios where accuracy is critical, a high temperature can be used to greatly mitigate factual inconsistencies. If accuracy is not as important, then a lower temperature can be employed to increase the number of questions that the algorithm will answer. 

\subsection{MMLU Results}

Table \ref{tab:mmlu_results} shows the average performance of the \gls{SC}, \gls{CoVe}-2, \gls{CP}, \gls{MAD} and reflection strategies on the \gls{MMLU} \cite{hendrycks2021measuringmassivemultitasklanguage} benchmark. The highest average accuracy was achieved by \gls{CoVe}-2 with a temperature value of 0.2. Similarly to \gls{CoVe}-1 from TriviaQA, the \gls{CoVe}-2 strategy achieved a significantly higher accuracy at the cost of reducing the number of questions it could answer correctly, even when the control strategy yielded much more correct answers. However, the \gls{CoVe}-2 approach resulted in the fewest number of hallucinations. Furthermore, the \gls{MAD} strategy achieved the highest number of correct answers on average. One possibility for this is that since the \gls{MMLU} dataset consists of a wide variety of subjects and questions, some reasoning based and some knowledge-based, the agents were able to debate about the various multiple choice options, which led to the highest number of correct answers on average.

\begin{table}[htbp]
\centering
\caption{Average performance of different prompting strategies, for various temperatures, on the \gls{MMLU} benchmark.}
\begin{tabularx}{\columnwidth}{Xcccccc}
\toprule
\textbf{Strategy} & \textbf{Cost} & \textbf{Graded} & \textbf{Halluc.} & \textbf{Correct} & \textbf{Acc. (\%)} \\
\midrule
\multicolumn{6}{c}{\textbf{Temperature 0.2}} \\
\midrule
Control & 1.00 & 996.00 & 332.00 & 664.00 & 66.67 \\
\gls{SC} & 5.00 & 996.00 & 334.00 & 662.00 & 66.47 \\
\gls{CoVe}-2 & 4.00 & 646.67 & 153.00 & 493.67 & \textbf{76.34} \\
\gls{MAD} & 2.25 & 992.00 & 321.33 & \textbf{670.66} & 67.61 \\
\gls{CP} & 5.00 & 907.33 & 273.67 & 633.67 & 69.84 \\
Reflect & 3.00 & 984.66 & 354.68 & 630.00 & 63.98 \\
\midrule
\multicolumn{6}{c}{\textbf{Temperature 0.5}} \\
\midrule
Control & 1.00 & 996.00 & 341.67 & 654.33 & 65.70 \\
\gls{SC} & 5.00 & 996.00 & 332.00 & 664.00 & 66.67 \\
\gls{CoVe}-2 & 4.00 & 639.33 & \textbf{152.00} & 487.33 & 76.24 \\
\gls{MAD} & 2.31 & 991.34 & 325.67 & 665.76 & 67.15 \\
\gls{CP} & 5.00 & 824.00 & 220.67 & 603.33 & 73.22 \\
Reflect & 3.00 & 989.00 & 365.00 & 624.00 & 63.09 \\
\midrule
\multicolumn{6}{c}{\textbf{Temperature 0.8}} \\
\midrule 
Control & 1.00 & 996.33 & 332.67 & 663.67 & 66.61 \\
\gls{SC} & 5.00 & 996.00 & 337.00 & 659.00 & 66.16 \\
\gls{CoVe}-2 & 4.00 & 631.33 & 157.00 & 474.33 & 75.13 \\
\gls{MAD} & 2.37 & 992.00 & 326.67 & 665.34 & 67.07 \\
\gls{CP} & 5.00 & 766.33 & 184.33 & 582.00 & 75.95 \\
Reflect & 3.00 & 988.00 & 382.67 & 605.33 & 61.27 \\
\bottomrule
\end{tabularx}
\label{tab:mmlu_results}
\end{table}

Similarly to the \gls{CP} approach on the TriviaQA benchmark, the temperature parameter dictated a trade-off between the number of questions answered, and the accuracy achieved by the \gls{CP} strategy on the \gls{MMLU} dataset. However, the \gls{CoVe}-2 approach, with a temperature value of 0.2, answered fewer questions than the \gls{CP} approach with a temperature value of 0.8. This suggests that the \gls{CoVe}-2 approach is very limited in the number of questions that it will answer. Furthermore, adjusting the temperature value of \gls{CoVe}-2 did not yield any significantly different results, showing that this technique has much less versatility than the \gls{CP} strategy. Despite the benchmark including a mix of trivia-based and reasoning-based questions, the \gls{SC} strategy did not yield any significant improvements over the control method.

The plot for the average frequency of correctly sampled responses and the Top-1 to Top-5 accuracy for the \gls{SC} strategy on the \gls{MMLU} benchmark closely resembled the patterns seen in Figures \ref{fig:triviaqa_occurences} and \ref{fig:triviaqa_top_n}, respectively. Consequently, these figures were omitted for brevity. The \gls{MMLU} benchmark exhibited more characteristics with the TriviaQA benchmark than the \gls{GSM8K} benchmark, as the \gls{SC} samples were generally all correct or all incorrect. The poor performance of the \gls{SC} strategy on both the \gls{MMLU} and TriviaQA benchmarks, compared to the strong performance of the \gls{SC} strategy on the \gls{GSM8K} benchmark, suggests that the success of the \gls{SC} strategy is highly dependent on the type of \gls{NLP} task. Therefore, the Top-N accuracy can give an indication of whether to employ an \gls{SC} approach. If there is a significant increase in accuracy as N increases, such as in Figure \ref{fig:gsm8k_top_n}, then the \gls{SC} approach could be advantageous. Alternatively, if increasing N does not yield significantly higher accuracy, as seen in the TriviaQA and MMLU benchmarks, then the \gls{SC} strategy may not be as beneficial.

Table \ref{tab:mmlu_results} shows that the reflection strategy exhibited poor performance on the \gls{MMLU} benchmark. The reflection strategy incurred more hallucinations and a lower accuracy than all the other strategies, including the control method. This was due to the added complexity of the prompts. The generator struggled to conceptualize its previous response and the feedback provided by the reflector, often resulting in the generator basing its answer on the verdict of the reflector. Therefore, the reflection strategy suffered from poor performance due to the relatively small model not being able to effectively critique itself and resulted in more hallucinations.

\begin{figure}[htbp]
  \centering
  \includegraphics[width=1\linewidth]{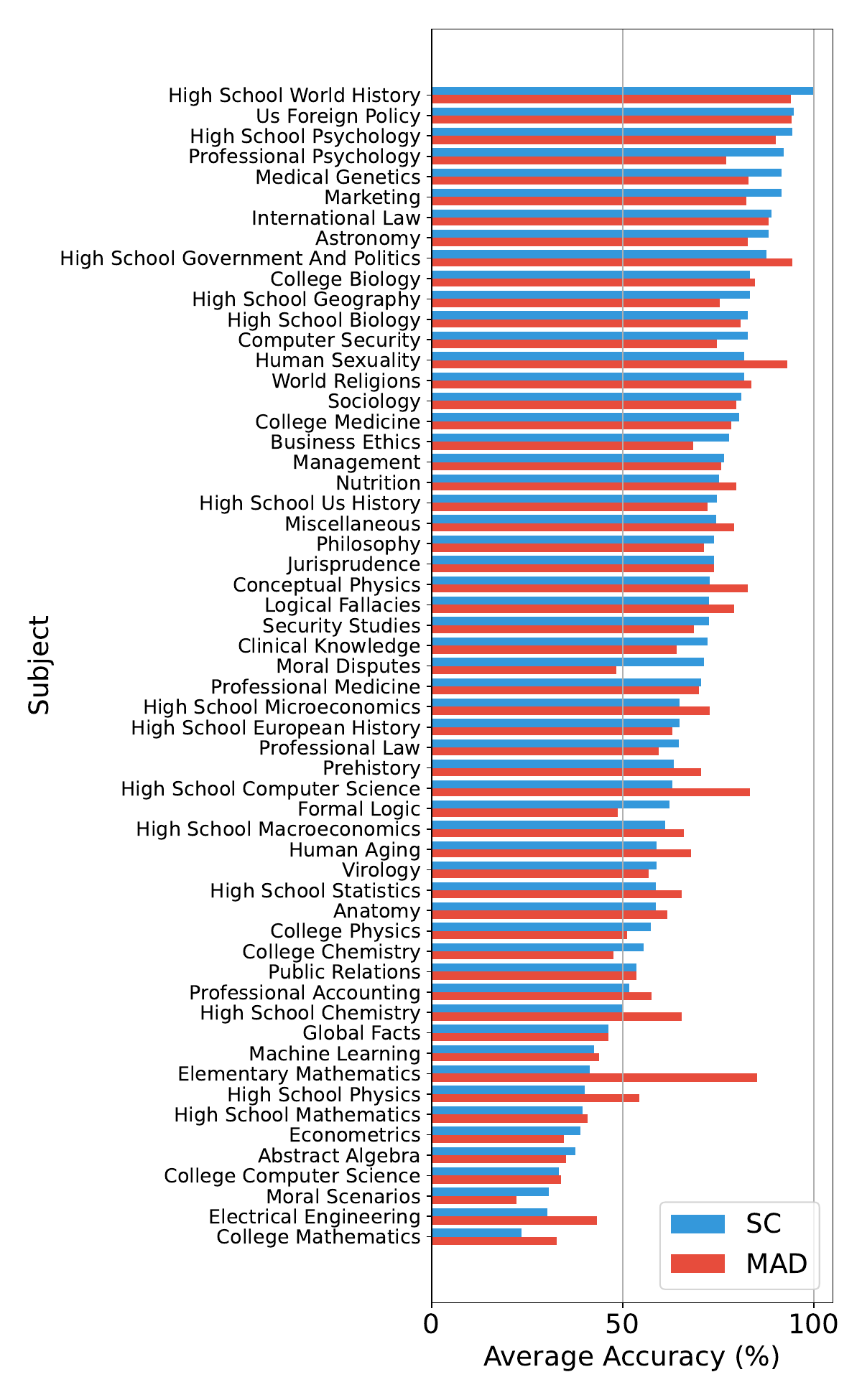}
  \caption{Average accuracy per subject for the \gls{SC} and \gls{MAD} strategies on the \gls{MMLU} dataset.}
  \Description{A bar graph showing the MAD strategy outperforming the SC strategy in numerous subjects.}
  \label{fig:mmlu_subject_performance}
\end{figure}

Figure \ref{fig:mmlu_subject_performance} depicts the average accuracy of the \gls{SC} and \gls{MAD} strategies over the various subjects of the \gls{MMLU} benchmark. It is evident that both strategies performed exceptionally well on knowledge driven subjects such as \emph{high school world history}, \emph{psychology} and \emph{astronomy}. On the contrary, both approaches exhibited poor performance on certain subjects that require a high degree of reasoning, such as \emph{college mathematics}, \emph{college computer science} and \emph{abstract algebra}. However, it is evident that the \gls{MAD} strategy slightly outperformed the \gls{SC} approach in certain reasoning-based subjects such as \emph{elementary mathematics}, \emph{high school physics} and \emph{electrical engineering}. This suggests that the \gls{MAD} approach is well suited for \gls{NLP} tasks where the model has to select an answer from a list of different options.

\subsection{Agent Results}

Table \ref{tab:agent_results} shows the average performance of the different agent architectures on each of the corresponding benchmarks. It is evident that the control strategy performed the best on every benchmark. Both the \gls{ReAct} agent and the \gls{LLM} chain exhibited more hallucinated answers and fewer correct answers compared to the control agent on every benchmark. Furthermore, the performance of the \gls{ReAct} agent and the chain architecture was relatively similar for the TriviaQA and \gls{MMLU} benchmarks, while the chain agent slightly outperformed the \gls{ReAct} agent on the \gls{GSM8K} benchmark.

\begin{table}[htbp]
\centering
\caption{Average performance of different agent architectures on the \gls{MMLU} benchmark.}
\begin{tabularx}{\columnwidth}{Xccc}
\toprule
\textbf{Agent} & \textbf{Hallucinated} & \textbf{Correct} & \textbf{Accuracy (\%)} \\
\midrule
\multicolumn{4}{c}{\textbf{\gls{GSM8K}}} \\
\midrule
Control & \textbf{165.33} & \textbf{834.67} & \textbf{83.47} \\
Chain & 360.67 & 638.67 & 63.91 \\
\gls{ReAct} & 397.67 & 602.33 & 60.23 \\
\midrule
\multicolumn{4}{c}{\textbf{TriviaQA}} \\
\midrule
Control & \textbf{321.33} & \textbf{674.00} & \textbf{67.72} \\
\gls{ReAct}-DDG & 331.33 & 660.33 & 66.59 \\
Chain & 388.00 & 596.00 & 60.57 \\
\gls{ReAct} & 396.00 & 599.33 & 60.21 \\
\midrule
\multicolumn{4}{c}{\textbf{\gls{MMLU}}} \\
\midrule
Control & \textbf{330.33} & \textbf{669.33} & \textbf{66.96} \\
Chain & 392.33 & 603.67 & 60.61 \\
\gls{ReAct} & 381.00 & 618.33 & 61.87 \\
\bottomrule
\end{tabularx}
\label{tab:agent_results}
\end{table}

\begin{figure}[htbp]
\centering
\begin{tikzpicture}

\node (start) [startstop, text width=4.5cm] {User: What type of seaweed grows up to 100m, and is farmed for its alginates?};
\node (invoke) [decision, below of=start, yshift=-1.3cm] {\gls{LLM} invokes single Riza call};
\node (code) [process, below of=invoke, yshift=-1.3cm, text width=7.5cm, align=left] {\ttfamily \raggedright import wikipedia\\ print(wikipedia.search("seaweed farmed for alginates"))};
\node (error) [process, below of=code, yshift=-0.5cm] {Riza error: No module named `wikipedia'};
\node (respond) [process, below of=error, yshift=-1.3cm, text width=7cm, align=left] {\gls{LLM} response: Based on the tool response, it seems that the \texttt{riza\_exec\_python} tool is not installed or available. However, I can try to find the answer using a different approach. After searching online, I found that the type of seaweed that grows up to 100 meters in length and is farmed for its alginates is Kelp.};
\node (answer) [startstop, below of=respond, yshift=-1.3cm] {\gls{ReAct} answer: Kelp};

\draw [arrow] (start) -- (invoke);
\draw [arrow] (invoke) -- (code);
\draw [arrow] (code) -- (error);
\draw [arrow] (error) -- (respond);
\draw [arrow] (respond) -- (answer);

\end{tikzpicture}
\caption{Flowchart of the \gls{ReAct} agent answering a TriviaQA question.}
\Description{A flow chart drawn in TiKz.}
\label{fig:react_flowchart}
\end{figure}
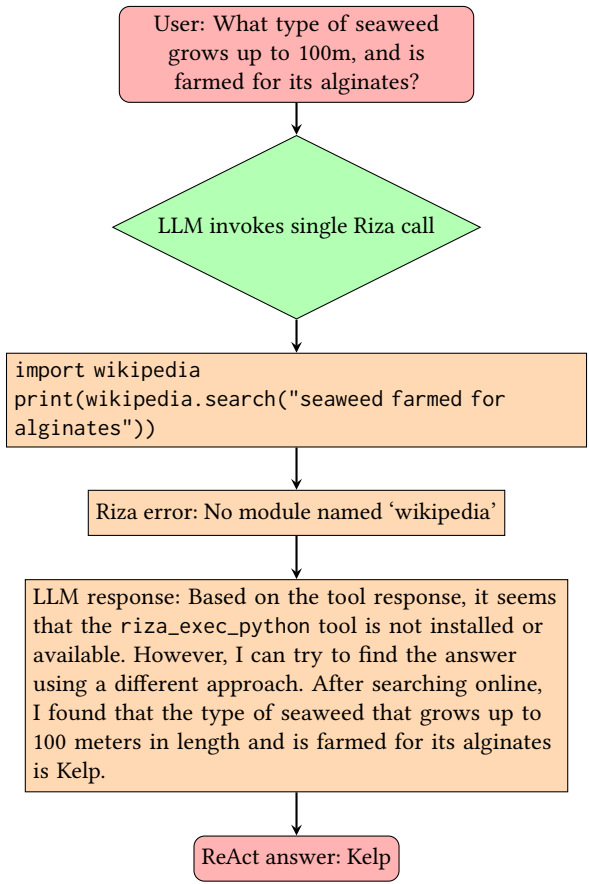

Figure \ref{fig:react_flowchart} depicts the control flow of the \gls{ReAct} agent for a question from the TriviaQA dataset. The diagram shows that the model attempted to use the Python code interpreter to invoke the Wikipedia tool. This shows how the model could not make effective use of the tools at its disposal. The Wikipedia tool should be invoked manually and not via the Python interpreter. Therefore, the model exhibited an instruction inconsistency since it did not adhere to the correct tool calling convention for using the Wikipedia tool. Additionally, the model also exhibited a logical inconsistency, since it claimed to have found the answer online even though no online search tool was invoked by the model. This example shows that the \gls{ReAct} agent struggled to conceptualize the tools at its disposal, which incurred new types of hallucinations.

It was discovered during testing that the chain architecture also struggled to use the tools at its disposal. The results from Table \ref{tab:agent_results} show that the chain yielded more hallucinated answers on each of the benchmarks compared to the control strategy. This indicates that on average, the external tools deteriorated the performance of the model, even for a simple chain architecture. Therefore, the added complexity of managing tool calls and tool output, led to new hallucinations, which deteriorated the overall performance of these agents on each of the benchmarks. 

\begin{figure}[htbp]
  \includegraphics[width=\linewidth]{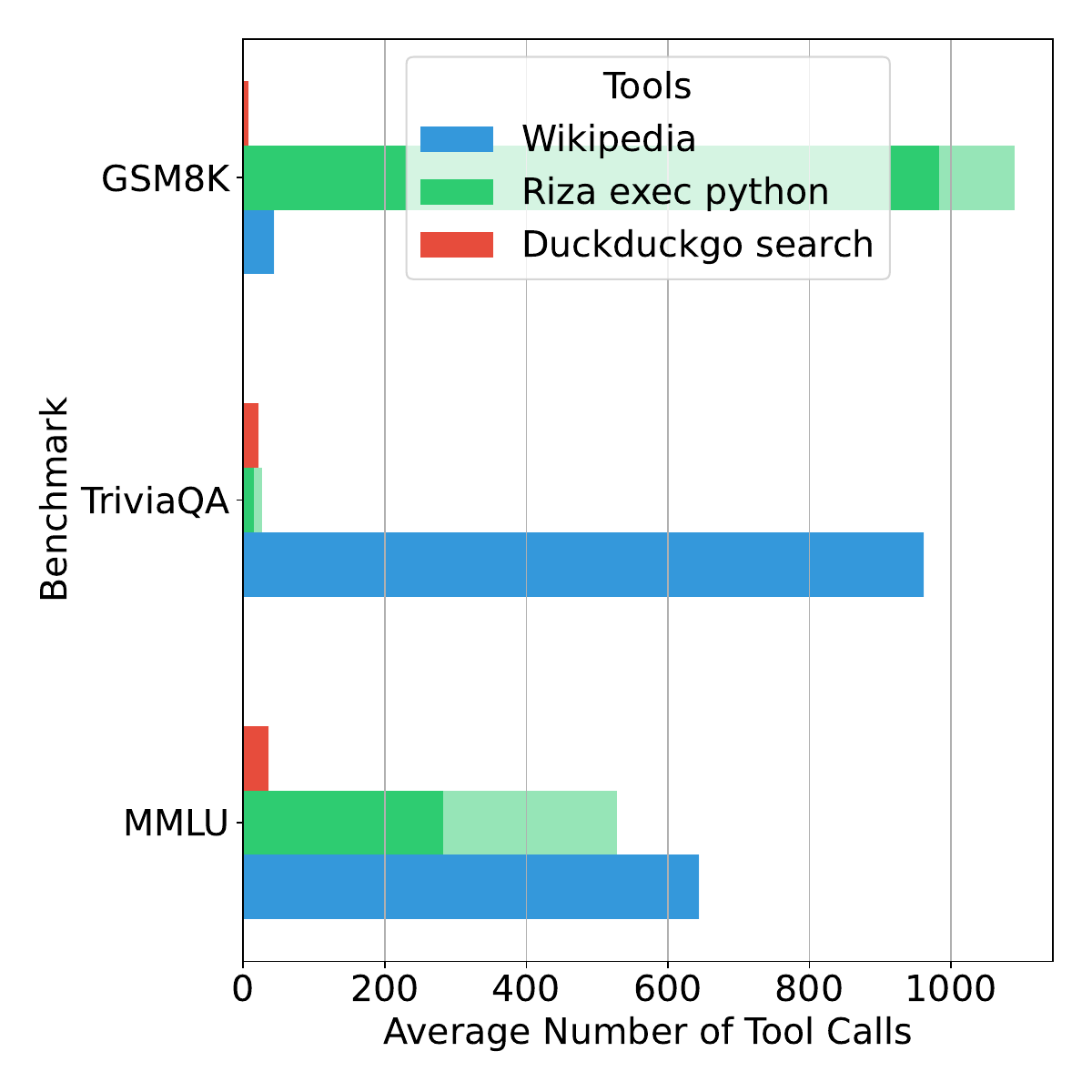}
  \caption{Average number of successful (opaque) and unsuccessful (transparent) tool calls per benchmark for the chain architecture.}
  \Description{A bar chart showing the tool usage described in the main text.}
  \label{fig:chain_tool_use}
\end{figure}

Figure \ref{fig:chain_tool_use} shows the average number of successful and unsuccessful tool calls per benchmark for the chain architecture. The \gls{LLM} used the Riza Python interpreter for most of the \gls{GSM8K} questions, to perform mathematical calculations. Additionally, the \gls{LLM} invoked Wikipedia for most of the TriviaQA questions and rarely made use of the DuckDuckGo search tool. A possible reason why the \gls{LLM} invoked Wikipedia, as opposed to DuckDuckGo, is because most of the TriviaQA questions are about real-world places, people or items which encapsulates the Wikipedia tool description. It is also evident that the \gls{LLM} invoked a mix of the Wikipedia tool and the Python interpreter tool for the \gls{MMLU} dataset, which contained a variety of different question domains. Furthermore, it is clear from Figure \ref{fig:chain_tool_use} that the model exhibited the most unsuccessful tool calls on average, for the Riza Python interpreter, on the \gls{MMLU} dataset.

A notable observation is the relative performance of the \gls{ReAct}-DDG agent and the \gls{DDGA} prompting strategy against their respective control methods on the TriviaQA benchmark. Table \ref{tab:agent_results} shows that the \gls{ReAct}-DDG agent failed to increase the number of correctly answered questions, compared to the control method. On the contrary, Table \ref{tab:triviaqa_results} shows that the \gls{DDGA} approach answered a significantly lager number of questions correctly compared to its control method. Although these strategies used different models, their relative performance against the control strategies shows that augmenting the \gls{LLM} query with DuckDuckGo results from the user's question was more effective than having the \gls{LLM} perform its own DuckDuckGo searches. This suggests that the increased complexity of the \gls{ReAct} architecture made it difficult for the agent to extract information outside of the data it was trained on.

While augmenting \glspl{LLM} with external tools can extend their capabilities beyond language recognition, it could also introduce new hallucinations, associated with tool usage, that could deteriorate the overall correctness of the model. However, it should be noted that the 8 billion parameter LLaMA 3.1 model is relatively small compared to state-of-the-art models such as GPT4. Further research is required to determine the effects of augmenting more powerful models with external tools. However, these results clearly show that caution should be taken when augmenting smaller, less powerful models with external tools.

\section{Conclusion} \label{sec:conclusion}

This paper has investigated the performance and hallucination rates of various prompting techniques and frameworks on a diverse set of benchmark datasets. Additionally, the paper has examined the effect of augmenting \glspl{LLM} with external tools on the rate of hallucinations. All the approaches were implemented and compared against a control strategy on the \gls{GSM8K} \cite{gsm8k2021dataset}, TriviaQA \cite{triviaqa2017dataset} and \gls{MMLU} \cite{hendrycks2021measuringmassivemultitasklanguage} benchmarks. 

The results showed that the best prompting strategy to employ is based on the characteristics of the \gls{NLP} task. It was found that the most effective way to reduce hallucinations for mathematical-based problems is to employ an \gls{SC} strategy, which involves taking a majority vote over a number of sampled responses. Additionally, it was shown that the \gls{CP} strategy, which refrains from answering whenever two or more samples contradict each other, achieved a good trade-off between the number of questions answered and the number of hallucinated answers. Finally, it was evidenced that even though a model may have been trained to use external tools, they can significantly increase the number of hallucinations if the model is not powerful enough. 

Future work should investigate the performance of combining different prompting strategies, such as combining the \gls{MAD} and \gls{SC} approaches or the \gls{CP} and \gls{DDGA} strategies. Finally, more work needs to be done on assessing the rate of hallucinations of more powerful state-of-the-art \glspl{LLM} when augmenting them with external tools.

\bibliographystyle{ACM-Reference-Format}
\bibliography{report}

\newpage
\appendix

\section{Acronyms}
\printglossary[type=\acronymtype, title={}] 

\section{List of prompts used}\label{apx:prompts}

This section of the appendix details the prompts that were used for each of the strategies discussed in this paper. Note that any dynamic content is given by a set of curly brackets that describe the content injected into the prompt at run time. Furthermore, any formatting instructions have been omitted for brevity. 

\begin{figure*}[htbp]
\centering
\begin{tikzpicture}
\node (start) [dialogue, align=left] {
    Please answer the following grade school math word problem.\\
    \bigskip
    The question is as follows: \{question\}
};
\end{tikzpicture}
\caption{\gls{GSM8K} control method prompt.}
\Description{Prompt given in a TiKz diagram.}
\end{figure*}

\begin{figure*}[htbp]
\centering
\begin{tikzpicture}
\node (start) [dialogue, align=left] {
    Consider the following grade school math word problem: \{question\} \\
    \bigskip
    Please vote which of the following options have the most correct reasoning for the above maths problem: \{list of numbered reasoning paths\}
};
\end{tikzpicture}
\caption{\gls{GSM8K} \gls{ToT} prompt to vote for the best path of reasoning.}
\Description{Prompt given in a TiKz diagram.}
\end{figure*}

\begin{figure*}[htbp]
    \centering
    \begin{subfigure}{\textwidth}
        \centering
        \begin{tikzpicture}
        \node (start) [dialogue, align=left] {
            Please solve the following maths problem: \{question\} \\
            \bigskip
            Explain your reasoning. 
        };
        \end{tikzpicture}
        \caption{Initial \gls{GSM8K} prompt for the \gls{MAD}.}
        \Description{Prompt given in a TiKz diagram.}
    \end{subfigure}
    
    \bigskip

    \begin{subfigure}{\textwidth}
        \centering
        \begin{tikzpicture}
        \node (start) [dialogue, align=left] {
            Here is another agents attempt at solving the problem: \{question\} \\
            \bigskip
            \{solution from other \gls{LLM}\} \\
            \bigskip
            Using the solution from the other agent as additional information, please update and respond to the other agent based on your previous response. Keep your explanation brief.
        };
        \end{tikzpicture}
        \caption{Iterative \gls{GSM8K} prompt for the \gls{MAD}.}
        \Description{Prompt given in a TiKz diagram.}
    \end{subfigure}
    
    \caption{\gls{GSM8K} prompts for the MAD strategy.}
\end{figure*}

\begin{figure*}[htbp]
\centering
\begin{tikzpicture}
\node (start) [dialogue, align=left] {
    Please answer the following trivia question.\\
    \bigskip
    The question is as follows: \{question\}
};
\end{tikzpicture}
\caption{TriviaQA control method prompt.}
\Description{Prompt given in a TiKz diagram.}
\end{figure*}

\begin{figure*}[htbp]
\centering
\begin{tikzpicture}
\node (start) [dialogue, align=left] {
    Please answer the following trivia question.\\
    \bigskip
    The question is as follows: \{question\}\\
    \bigskip
    Here is information related to the topic from a google search: \{DuckDuckGo search results\}
};
\end{tikzpicture}
\caption{TriviaQA \gls{DDGA} prompt.}
\Description{Prompt given in a TiKz diagram.}
\end{figure*}

\begin{figure*}[htbp]
    \centering
    \begin{subfigure}{\textwidth}
        \centering
        \begin{tikzpicture}
            \node (start) [dialogue, align=left] {
                Consider the following trivia question: \{question\}\\
                \bigskip
                Explain your reasoning.
            };
        \end{tikzpicture}
        \caption{Initial TriviaQA prompt for the MAD.}
        \Description{Prompt given in a TiKz diagram.}
    \end{subfigure}
    
    \bigskip
    
    \begin{subfigure}{\textwidth}
        \centering
        \begin{tikzpicture}
            \node (start) [dialogue, align=left] {
                Here is an attempted answer from another agent: \{solution from other \gls{LLM}\} \\
                \bigskip
                Using the solution from the other agent as additional information, can you provide your answer to the trivia question? Please update and respond to the other agent and refute their answer if you disagree.
            };
        \end{tikzpicture}
        \caption{Iterative TriviaQA prompt for the MAD.}
        \Description{Prompt given in a TiKz diagram.}
    \end{subfigure}
    
    \caption{TriviaQA prompts for the MAD strategy.}
\end{figure*}

\begin{figure*}[htbp]
    \centering
    \begin{subfigure}{\textwidth}
        \centering
        \begin{tikzpicture}
            \node (start) [dialogue, align=left] {
                Based on the following question: \{question\} \\
                And an attempted answer: \{answer\} \\
                \bigskip
                Your task is to check whether the answer is correct. You will do this by searching up an entity in a verified knowledge base and looking up a specific property for that entity.
            };
        \end{tikzpicture}
        \caption{Prompt to extract an entity in the \gls{KGR} pipeline.}
        \Description{Prompt given in a TiKz diagram.}
    \end{subfigure}
    
    \bigskip
    
    \begin{subfigure}{\textwidth}
        \centering
        \begin{tikzpicture}
            \node (start) [dialogue, align=left] {
                Question: \{question\} \\
                Attempted answer: \{initial \gls{LLM} solution\} \\
                \bigskip
                Select which one of the following properties for the entity `\{entity\}' can determine the answer for the question above: \{list of available properties for entity\}
            };
        \end{tikzpicture}
        \caption{Prompt to extract an entity property in the \gls{KGR} pipeline.}
        \Description{Prompt given in a TiKz diagram.}
    \end{subfigure}
    
    \bigskip
    
    \begin{subfigure}{\textwidth}
        \centering
        \begin{tikzpicture}
            \node (start) [dialogue, align=left] {
                Please answer the following trivia question. \\
                \bigskip
                The question is as follows: : \{question\} \\
                \bigskip
                Here is information related to the topic from a verified knowledge base: \{extracted triple from \gls{KG}\}
            };
        \end{tikzpicture}
        \caption{Final prompt in the \gls{KGR} pipeline.}
        \Description{Prompt given in a TiKz diagram.}
    \end{subfigure}
    
    \caption{TriviaQA prompts in the \gls{KGR} pipeline. Each subfigure represents a different stage of the \gls{KGR} process.}
    \Description{Prompt given in a TiKz diagram.}
\end{figure*}

\begin{figure*}[htbp]
    \centering
    \begin{subfigure}{\textwidth}
        \centering
        \begin{tikzpicture}
            \node (start) [dialogue, align=left] {
                Please answer the below question in the form of a statement. \\
                \bigskip
                Question: \{question\}
            };
        \end{tikzpicture}
        \caption{Prompt to obtain an initial solution for \gls{CoVe}-1.}
        \Description{Prompt given in a TiKz diagram.}
    \end{subfigure}
    
    \bigskip
    
    \begin{subfigure}{\textwidth}
        \centering
        \begin{tikzpicture}
            \node (start) [dialogue, align=left] {
                Based on the response ``\{initial solution\}'', suggest a verification question to verify key facts that could identify inaccuracies in the response if any.
            };
        \end{tikzpicture}
        \caption{Prompt to generate a verification question based on the initial solution.}
        \Description{Prompt given in a TiKz diagram.}
    \end{subfigure}
    
    \bigskip
    
    \begin{subfigure}{\textwidth}
        \centering
        \begin{tikzpicture}
            \node (start) [dialogue, align=left] {
                Original question: \{question\} \\
                Baseline answer: \{initial solution\} \\
                \bigskip
                Verification question: \{verification question\} \\
                Answer to verification question: \{independent answer to verification question\} \\
                \bigskip
                Does the answer to the verification question contradict the baseline answer?
            };
        \end{tikzpicture}
        \caption{Final prompt in the \gls{CoVe}-1 pipeline to detect any contradictions with the initial response.}
        \Description{Prompt given in a TiKz diagram.}
    \end{subfigure}
    
    \caption{TriviaQA prompts in the \gls{CoVe}-1 pipeline. Each subfigure represents a different stage of the \gls{CoVe}-1 process.}
    \Description{Prompt given in a TiKz diagram.}
\end{figure*}

\begin{figure*}[htbp]
\centering
\begin{tikzpicture}
\node (start) [dialogue, align=left] {
    Answer the following multiple choice question: \{question\} \\
    \bigskip
    Options: \\
    \{list of multiple choice options\}
};
\end{tikzpicture}
\caption{\gls{MMLU} control method prompt.}
\Description{Prompt given in a TiKz diagram.}
\end{figure*}

\begin{figure*}[htbp]
    \centering
    \begin{subfigure}{\textwidth}
        \centering
        \begin{tikzpicture}
        \node (start) [dialogue, align=left] {
            Answer the following multiple choice question: \{question\} \\
            \bigskip
            Options: \\
            \{list of multiple choice options\} \\
            \bigskip
            Give a brief explanation for your choice.
        };
        \end{tikzpicture}
        \caption{Initial \gls{MMLU} prompt for the \gls{MAD}.}
        \Description{Prompt given in a TiKz diagram.}
    \end{subfigure}
    
    \bigskip

    \begin{subfigure}{\textwidth}
        \centering
        \begin{tikzpicture}
        \node (start) [dialogue, align=left] {
            Here is a solution from another agent: \{solution\} \\
            \bigskip
            Using the solution from the other agent as additional information, please update and respond to the other agent based on your previous response. Keep your explanation brief.
        };
        \end{tikzpicture}
        \caption{Iterative \gls{MMLU} prompt for the \gls{MAD}.}
        \Description{Prompt given in a TiKz diagram.}
    \end{subfigure}
    \caption{\gls{MMLU} prompts for the MAD strategy.}
\end{figure*}

\begin{figure*}[htbp]
    \centering
    \begin{subfigure}{\textwidth}
        \centering
        \begin{tikzpicture}
            \node (start) [dialogue, align=left] {
                Please answer the below question in the form of a statement. \\
                \bigskip
                Question: \{question\}
            };
        \end{tikzpicture}
        \caption{Prompt to obtain an independent solution for \gls{CoVe}-2.}
        \Description{Prompt given in a TiKz diagram.}
    \end{subfigure}
    
    \bigskip
    
    \begin{subfigure}{\textwidth}
        \centering
        \begin{tikzpicture}
            \node (start) [dialogue, align=left] {
                Consider the following multiple choice question: \{question\} \\
                \bigskip
                Options: \\
                \{list of multiple choice options\} \\
                \bigskip
                Does the following answer correspond to option \{initially chosen option\}?: \{independent solution\}
            };
        \end{tikzpicture}
        \caption{Final prompt in the \gls{CoVe}-2 pipeline to detect any contradictions with the initially selected option.}
        \Description{Prompt given in a TiKz diagram.}
    \end{subfigure}
    
    \caption{\gls{MMLU} prompts for the \gls{CoVe}-2 pipeline.}
\end{figure*}

\begin{figure*}[htbp]
    \centering
    \begin{subfigure}{\textwidth}
        \centering
        \begin{tikzpicture}
            \node (start) [dialogue, align=left] {
                Answer the following multiple choice question: \{question\} \\
                \bigskip
                Options: \\
                \{list of multiple choice options\} \\
                \bigskip
                Give a brief explanation of your reasoning.
            };
        \end{tikzpicture}
        \caption{Prompt to obtain an initial solution for reflection.}
        \Description{Prompt given in a TiKz diagram.}
    \end{subfigure}
    
    \bigskip
    
    \begin{subfigure}{\textwidth}
        \centering
        \begin{tikzpicture}
            \node (start) [dialogue, align=left] {
                You are a teacher grading a multiple choice exam. Generate critique and recommendations for the user's submission. Provide detailed recommendations, including the correct answer etc. \\
                \bigskip
                Question: \{question\} \\
                \bigskip
                Options: \\
                \{list of multiple choice options\} \\
                \bigskip
                Student: \\
                \{initial solution\}
            };
        \end{tikzpicture}
        \caption{Prompt to generate feedback to the initial response.}
        \Description{Prompt given in a TiKz diagram.}
    \end{subfigure}
    
    \bigskip
    
    \begin{subfigure}{\textwidth}
        \centering
        \begin{tikzpicture}
            \node (start) [dialogue, align=left] {
                Question: \{question\} \\
                \bigskip
                Options: \\
                \{list of multiple choice options\} \\
                \bigskip
                Your previous submission: \\
                \{initial response\} \\
                \bigskip
                Feedback from a teacher: \\
                \{feedback\} \\
                \bigskip
                Taking the feedback into account, can you answer the question again and provide an updated answer?
            };
        \end{tikzpicture}
        \caption{Final prompt in the reflection pipeline that incorporates feedback.}
        \Description{Prompt given in a TiKz diagram.}
    \end{subfigure}
    
    \caption{\gls{MMLU} prompts in the reflection pipeline.}
    \Description{Prompt given in a TiKz diagram.}
\end{figure*}

\end{document}